\documentclass[letterpaper, 10 pt, conference]{ieeeconf}  

\IEEEoverridecommandlockouts                              

\usepackage{amsmath}
\usepackage{nccmath}
\usepackage{algorithmic}
\usepackage{algorithm}
\usepackage{amsfonts}
\usepackage{caption}
\usepackage{amsmath}
\usepackage{makecell}
\usepackage{multirow}
\usepackage{multicol}
\usepackage{caption}
\usepackage{subcaption}
\usepackage{graphicx}
\usepackage{verbatim}
\usepackage{amsmath,amssymb}
\usepackage{mathtools}
\usepackage{amsmath,mleftright}
\usepackage{xparse}
\usepackage{tikz}
\usepackage{graphicx}
\usepackage{pdfrender}
\usetikzlibrary{positioning}

\NewDocumentCommand{\evalat}{sO{\big}mm}{%
  \IfBooleanTF{#1}
   {\mleft. #3 \mright|_{#4}}
   {#3#2|_{#4}}%
}

\title{\LARGE \bf
Confidence Guided Stereo 3D Object Detection with Split Depth Estimation}

\author{Chengyao Li, Jason Ku, and Steven L. Waslander
\thanks{Authors are with the University of Toronto Institute for Aerospace Studies.
        {email: (chengyao.li@mail.utoronto.ca)}, {(kujason.ku@mail.utoronto.ca)}, {(stevenw@utias.utoronto.ca)}}%
}

\begin{document}

\maketitle
\thispagestyle{empty}
\pagestyle{empty}

\begin{abstract}
Accurate and reliable 3D object detection is vital to safe autonomous driving. Despite recent developments, the performance gap between stereo-based methods and LiDAR-based methods is still considerable. Accurate depth estimation is crucial to the performance of stereo-based 3D object detection methods, particularly for those pixels associated with objects in the foreground. Moreover, stereo-based methods suffer from high variance in the depth estimation accuracy, which is often not considered in the object detection pipeline. To tackle these two issues, we propose CG-Stereo, a confidence-guided stereo 3D object detection pipeline that uses separate decoders for foreground and background pixels during depth estimation, and leverages the confidence estimation from the depth estimation network as a soft attention mechanism in the 3D object detector. Our approach outperforms all state-of-the-art stereo-based 3D detectors on the KITTI benchmark.
\end{abstract}

\section{INTRODUCTION}
3D object detection is vital to applications such as autonomous driving. Many LiDAR-based methods~\cite{ku2018joint, chen2017multi, shi2019pointrcnn} achieve strong performance due to the accurate depth information that LiDAR provides. Compared with LiDAR, a stereo camera setup is less expensive and provides more dense information. In addition, stereo detection could add redundancy to an autonomous driving system and help reduce safety risks in combination with LiDAR methods. Recent stereo-based methods~\cite{wang2019pseudo, you2019pseudo, pon2019object, li2019stereo} have shown promising performance, although the detection performance gap between stereo and LiDAR configurations is still considerable. 

\begin{figure}[t] 
	\begin{center}
		\includegraphics[width=1\linewidth]{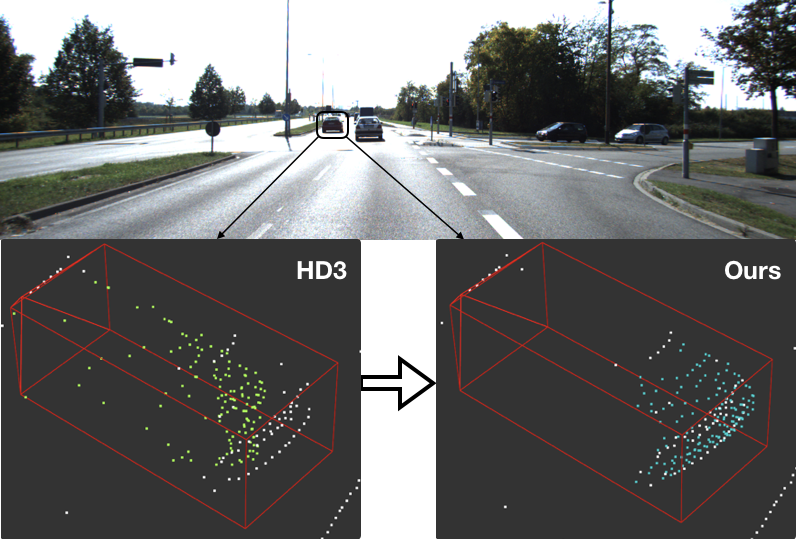}
	\end{center}
		\caption{A comparison between our proposed depth estimation module with our baseline HD$^3$ on KITTI dataset. Using LIDAR measurements (shown in white) as a reference, our proposed method (shown in dark cyan) is able to learn the depth and shape of the car more accurately compared with our baseline (shown in yellow). The 3D bounding box is shown in red.}
		\label{fig:estimated_pc}
\end{figure}

One recent state-of-the-art stereo-based approach is Pseudo-LiDAR~\cite{wang2019pseudo, you2019pseudo}, which first estimates disparities with a stereo matching network, converts the estimated disparities into a 3D point cloud, and then feeds the estimated point cloud to a LiDAR-based 3D object detector. However, compared with the LiDAR point cloud, the estimated point cloud often has poor depth estimation, particularly as depth increases, where it no longer preserves the shape of the objects. One attribute of this method is that the stereo matching algorithm jointly estimates both foreground and background pixels and does not learn specifically the depth and shape of the foreground objects.~\cite{wang2019task} shows that the depth distribution and pattern for foreground pixels and background pixels are different, and treating foreground and background pixels equally leads to sub-optimal results in a monocular depth estimation pipeline. In this paper, we propose to use two separate decoders for foreground and background pixels in a stereo matching network to provide better estimates of object shape and depth. The foreground and background masks can be obtained from image segmentation. In cooperation with the point cloud loss from~\cite{ku2019monocular}, we show in Fig.~\ref{fig:estimated_pc} that with our approach the depth and shape of the object can be significantly improved. With further experiments, we show that such improvement also leads to better 3D object detection performance. 

\begin{figure}[t] 
	\begin{center}
		\includegraphics[width=1\linewidth]{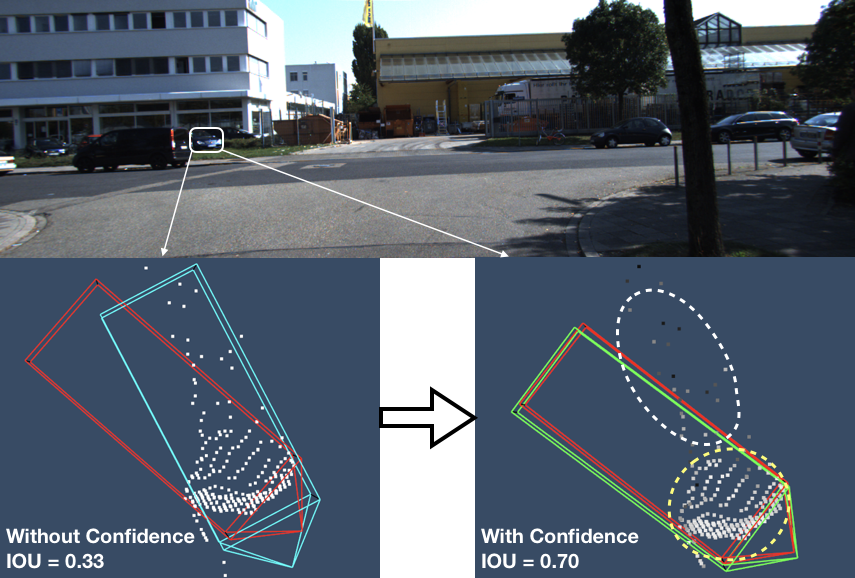}
	\end{center}
		\caption{A comparison between object bounding box detections without confidence (shown in teal) and with confidence (shown in green). Ground truth bounding box is shown in red. In the right image, points with higher confidence are shown in lighter colors. With confidence as an additional layer, the network is able to focus more on the points with high confidence (cycled in yellow) and ignore the points at object boundaries that have low confidence (cycled in white).  }
		\label{fig:confidence_visual}
\end{figure}

In contrast to LiDAR measured point clouds, the accuracy of stereo point clouds greatly varies across a scene. For constant pixel-level disparity error, the depth error increases as depth increases because of the effect of triangulation. In addition, the estimated point clouds also suffer from poor depth estimates at object boundaries, because it can be hard for a stereo matching algorithm to determine whether a pixel belongs to the object or the background~\cite{pon2019object}. In the original Pseudo-LiDAR pipeline~\cite{wang2019pseudo, you2019pseudo}, the estimated point cloud is directly fed into a 3D object detector which does not consider any uncertainty information. To address this issue, we propose to encode the uncertainty output from the depth estimation module as an additional layer in the point cloud serving as a soft attention mechanism.  This method allows the network to focus on points with high confidence and mitigates the effect of low confidence points such as points on the object boundary, as shown in Fig.~\ref{fig:confidence_visual}. 

In this paper, we propose CG-Stereo, a confidence guided stereo 3D object detection pipeline. To summarize, our main contributions are as follows:

\begin{itemize}
    \item We propose the use of separate depth decoders for foreground and background pixels in the stereo matching network, which leads to improved depth estimation accuracy of the foreground pixels and improved object detection performance.
    \item We propose the use of confidence estimates from the stereo matching algorithm as a soft attention mechanism to guide the object detector network to focus more on the points with higher quality depth information, leading to further improvement in object detection accuracy.
    \item  We demonstrate state-of-the-art performance that exceeds existing stereo-based methods on the challenging KITTI 3D object detection benchmark~\cite{geiger2012we} for all three object classes. Specifically, our approach surpasses the next best-performing method by 1.4\%, 6.7\%, and 12.7\% AP at 0.7 IOU on cars, pedestrians and cyclists, respectively.
\end{itemize}

\section{RELATED WORK}\label{related_work}
\noindent \textbf{Stereo depth estimation.}
For stereo vision, depth is often estimated by determining the stereo correspondences between the left and right images. Stereo matching is a well-established field of research~\cite{scharstein2002taxonomy}, with a long history of classical methods~\cite{tola2008fast, ohta1985stereo, kolmogorov2001computing}. With the recent development of deep learning, end-to-end learning methods have shown significant improvements in this task. A standard learning approach of stereo matching is to construct a 3D cost volume to minimize the matching cost~\cite{chang2018psmnet, zhang2019ganet, yin2019hd3}. Chang et al.~\cite{chang2018psmnet} propose a pyramid pooling module for incorporating global context followed by a stacked hourglass 3D CNN. Yin et al.~\cite{yin2019hd3} determine the stereo matching by decomposing the full match density into multiple scales hierarchically first and then compose a global match density. This method not only achieves state-of-the-art results on established benchmarks but also predicts a confidence map that indicates the certainty of the estimation for each pixel. Our stereo 3D detection method takes advantage of the confidence estimation and demonstrates they can be used to improve stereo 3D object detection performances. \\

\noindent \textbf{LiDAR-based 3D object detection.}
LiDAR-based object detection methods have shown strong performances and are widely used in autonomous driving since LiDAR provides accurate point clouds in terms of object depth and shape. Recent methods either use voxelization~\cite{chen2017multi, ku2018joint, zhou2018voxelnet, yan2018second}, PointNet~\cite{qi2018frustum, shi2019pointrcnn}, or a combination of the two~\cite{chen2019fast, shi2019pv,lang2019pointpillars} to learn features from point cloud data. Taking advantage of the mature pipeline of LiDAR-based detectors, the performance is transferable to stereo-based detectors.\\
\begin{figure*}[t!]
	\begin{center}
		\includegraphics[width=0.9\linewidth]{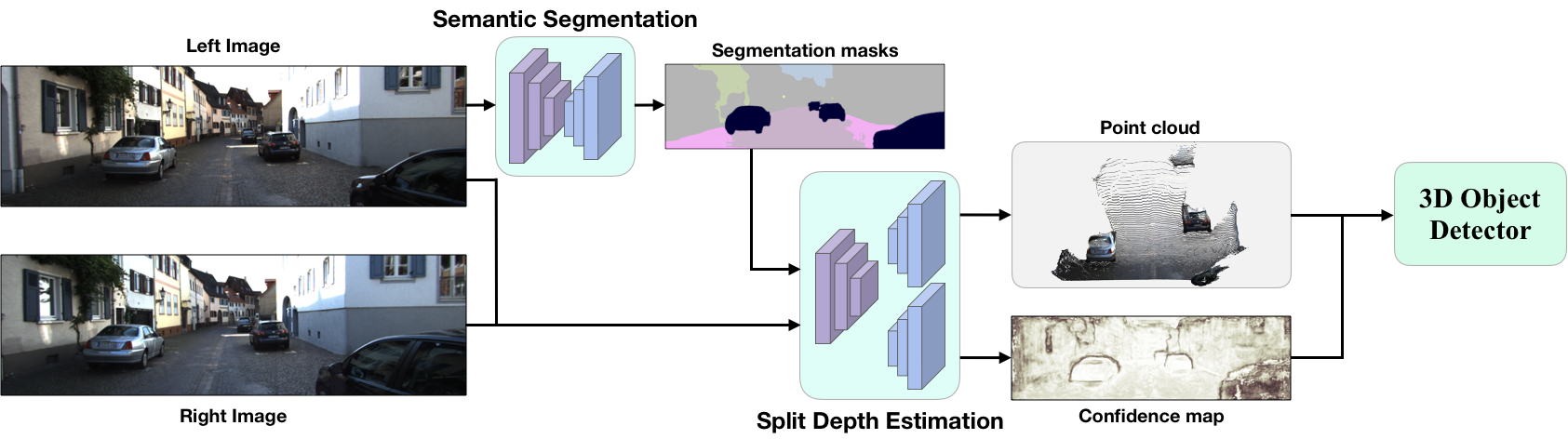}
	\end{center}
	\caption{Overview of our network. A semantic segmentation network first determines the foreground masks and background masks in the left image. The stereo matching network then estimates the disparities of the foreground and background pixels separately using two decoders and outputs the confidence associated with the estimation for each pixel. The disparity map is converted into a 3D point cloud with a confidence score as an additional layer.  Points belonging to the background are filtered out except for the ones belonging to the ground plane. The resulting point cloud is then fed into a point cloud-based 3D object detector.}
	\label{fig:architecture}
\end{figure*}

\noindent \textbf{Stereo-based 3D object detection.}
In recent years, stereo-based 3D object detection methods have shown promising improvements in performance. Stereo R-CNN~\cite{li2019stereo} combines 2D proposals from both left and right images along with the sparse keypoints to generate coarse 3D bounding boxes, and then refines the bounding box using the photometric alignment of left and right regions of interest (ROIs). TLNet~\cite{qin2019triangulation} projects the predefined anchor box to stereo images to obtain a pair of RoIs, learns to offset these RoIs, and uses triangulation to localize the objects. RT3DStereo~\cite{konigshof2019realtime} proposes to use semantic information together with disparity information to recover 3D bounding boxes. However, they do not take advantage of the semantic information to obtain better depth estimation. Pseudo-LiDAR~\cite{wang2019pseudo, you2019pseudo} proposes to mimic the LiDAR signal by converting the depth map to a point cloud, and then feed this point cloud to a LiDAR-based detector. This intuitive method reduces the performance gap between the stereo-based methods and LiDAR-based methods. However, the point cloud from stereo matching preserves streaking artifacts at the object boundaries, leading to inaccurate bounding box estimates. OC-Stereo~\cite{pon2019object} tries to solve this issue by estimating disparity only on the associated 2D bounding box area. However, this approach requires the 2D detection of the objects to be successful in both left and right images, which is difficult for objects that are truncated on image boundaries or are occluded from one view. It also completely ignores the background pixels which provide context to the 3D scene. Our method estimates the point cloud for both foreground and background pixels and keeps the points belonging to the ground plane since they contain useful contextual information in the 3D detection phase. The most recent state-of-the-art method, DSGN~\cite{chen2020dsgn}, proposes the use of a differentiable 3D volumetric representation of the environment to solve stereo 3D object detection. Their method achieves remarkable results on the KITTI car class, but the performances on pedestrians are not as competitive with the state-of-the-art. In comparison, our decomposed architecture allows us to perform well on pedestrians and cyclists even with the limited training data available for these classes in the KITTI dataset.    \\

\section{ARCHITECTURE}\label{Approach}
The overall pipeline of our method is shown in Fig.~\ref{fig:architecture}. First, a semantic segmentation network determines the foreground pixels and background pixels in the left image. We define foreground pixels as the pixels that belong to the objects of interest and background as all other pixels. Then, the stereo matching network estimates the disparities of the foreground and background pixels separately with two separate decoders. It also generates a confidence map associated with each pixel representing the certainty of the network's estimation. The disparity map is converted into a 3D point cloud with confidence as an additional layer. Points belonging to the background are filtered out except for the ones that lie on the ground plane. We use the same method as 3DOP~\cite{chen20153d} for stereo ground plane generation. The remaining point cloud is finally fed into a LiDAR-based 3D object detector.

\subsection{Semantic Segmentation}
For the 3D object detection task, it is common to segment the sensor input depending on the objects of interest. Pseudo-LiDAR~\cite{wang2019pseudo, you2019pseudo} converts the stereo image pair to a point cloud and then relies on a LiDAR-based 3D object detector to find objects. However, compared with LiDAR point clouds, the estimated point clouds have lower accuracy and thus are harder to segment. In addition, the texture and color information is lost in this process. We argue that it is possible to leverage image segmentation information in stereo object detection for improved detection accuracy, rather than relying on the LiDAR-based 3D detector exclusively. We also show that the foreground and background masks from semantic segmentation improve the depth estimation of the foreground pixels in Section~\ref{depth estimation}, which contributes to higher 3D object detection accuracy.

\subsection{Stereo Split Depth Estimation}\label{depth estimation} 
Our stereo depth estimation is performed via stereo matching and the proposed formulation is agnostic to any stereo matching algorithm. We build on top of HD$^3$ due to its state-of-the-art performance and ability to run in real-time. In addition, due to its probabilistic framework for match distribution estimation, an uncertainty associated with the estimate at every pixel can be naturally derived~\cite{yin2019hd3}. \\

\begin{figure*}[t!]
	\begin{center}
		\includegraphics[width=0.9\linewidth]{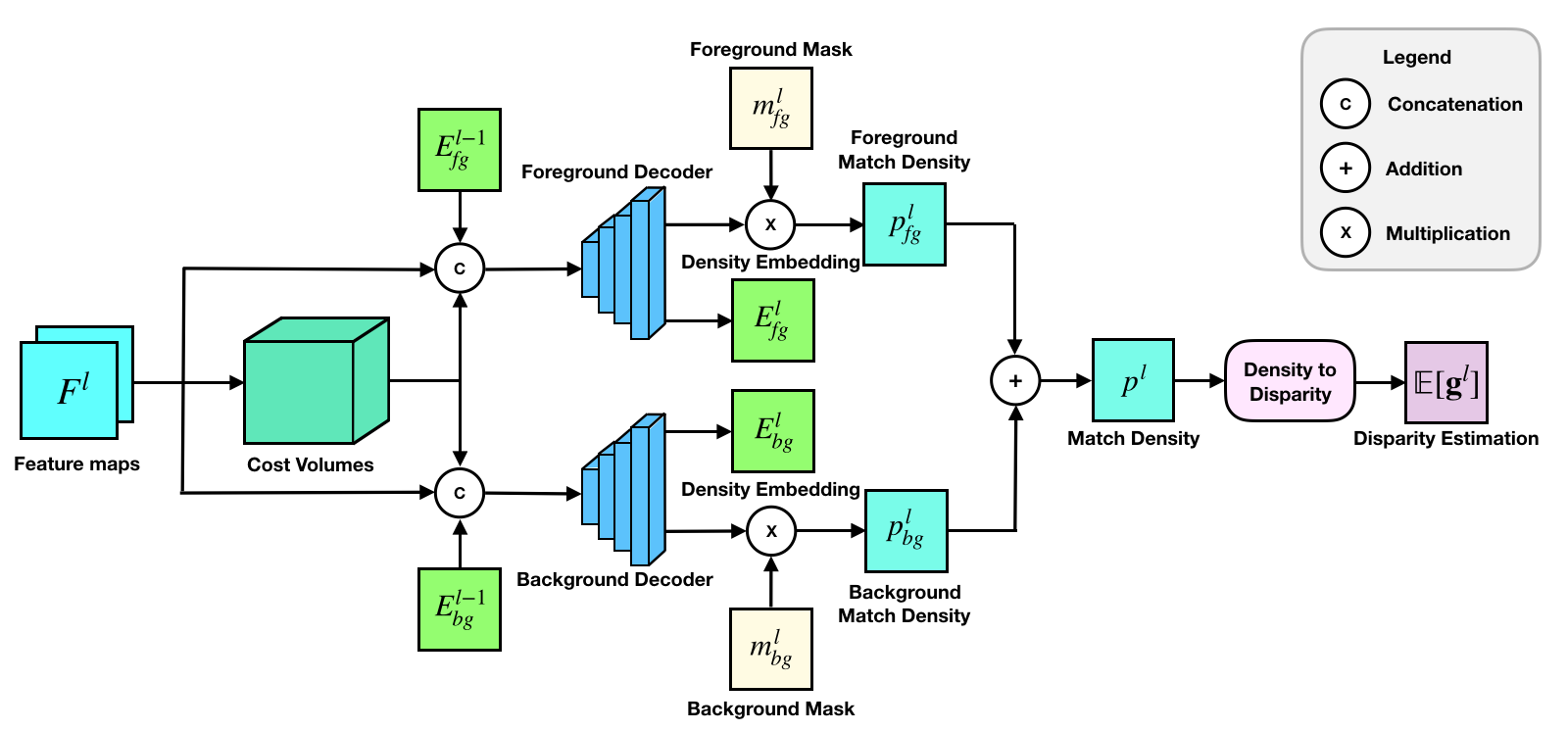}
	\end{center}
	\caption{Modified HD$^{3}$ network at the $l^{th}$ level. Instead of one density decoder for the entire image, we use separate density decoders for foreground pixels and background pixels, respectively, which allows us to optimize the weights specifically for each task.}
	\label{fig:hd3}
\end{figure*}

\noindent \textbf{Stereo Matching Architecture.} 
HD$^3$ is designed for learning probabilistic pixel correspondences in both optical flow and stereo matching tasks~\cite{yin2019hd3}, and we employ their stereo matching implementation as a baseline. The core idea of HD$^3$ is to decompose the full discrete match distributions of pixel-wise correspondences into multiple scales hierarchically, estimate the local matching distributions at each scale, and then compose them from all levels. The resulting distributions at each pixel in the reference image are referred to as match densities. At each image scale level $l$, a cost volume is constructed to find the correlation between the pixels in both images and a density decoder is trained to estimate the decomposed match density $p^l$. For more details, readers may refer to the original HD$^3$ paper~\cite{yin2019hd3}.\\

Our modified version of the HD$^3$ network is shown in Fig.~\ref{fig:hd3}. We only show one level for simplicity. Instead of relying on one density decoder for the entire image, we have two parallel density decoders with the same structure for foreground and background pixels, respectively. At each level $l$, each decoder takes the feature maps $F^{l}$, cost volume, and the density embedding from the previous
level $E^{l-1}_{fg}$ or $E^{l-1}_{bg}$ as input, and outputs an estimated match density $p^l_{fg}$ or $p^l_{bg}$, and the density embedding at the current level $E^{l}_{fg}$ or $E^{l}_{bg}$. Then, we use the foreground and background masks, denoted as $m^l_{fg}$ and $m^l_{bg}$, to mask out the output from the two decoders, and then fuse them. The match density is then converted into an estimated residual disparity $\mathbb{E}[\mathbf{g}^l]$ at the current level. The model-inherent uncertainty can be estimated by applying a \textit{softmax} operation and a \textit{max-pooling} operation to the estimated match density at the highest level.\\

\noindent \textbf{Loss Function.} 
We adapt the foreground-background sensitive loss function from a monocular-based method~\cite{wang2019task} and add the point cloud loss from~\cite{ku2019monocular}. The total loss is defined as
\begin{align}
   L_{total} = \lambda_f L_{fg} + (1-\lambda_f) L_{bg} + \alpha L_{pc}
\end{align}
where $\lambda_f$ is the weight coefficient representing the degree of preference for foreground pixels. $L_{fg}$ and $L_{bg}$ are the Kullback-Leibler divergence loss for foreground pixels and background pixels, respectively, $\alpha$ is a weight coefficient, and $L_{pc}$ is the point cloud loss, which is set to be a smooth L1 loss of the difference between the foreground point cloud $p_c$ with its ground truth point cloud $p_{gt}$ in camera frame. We include the loss for background pixels because there is interdependence between foreground and background pixels for inferring the depth~\cite{wang2019task}, and background provides context and support for 3D box regression~\cite{shi2019pointrcnn}. We include a point cloud loss because it directly penalizes the estimated 3D point cloud and further improves the 3D detection accuracy as demonstrated in Sec.~\ref{ablation}.

\subsection{Point Cloud Generation}
The disparity map is converted to 3D points using the camera projection model as shown in Eq.~\ref{depthtopoint}.
\begin{align}
    x = \frac{(u - c_u)z}{f_u}, 
    y = \frac{(v - c_v)z}{f_v}, 
    z =  \frac{f_ub}{d}
    \label{depthtopoint}
\end{align}
where $x$, $y$, $z$ is the position of the points, $(c_u, c_v)$ is the camera center, $(f_u, f_v)$ is the focal length, $b$ is the baseline, and $d$ is the estimated disparity for a given pixel.

Image segmentation is a relatively mature field and can provide robust results. As a consequence, rather than feeding the entire point cloud to the 3D object detector, we leverage the segmentation masks and feed only the points that are estimated to be from the foreground pixels. We also keep the points that belong to the ground plane as they provide useful contextual information and supporting information for generating proposals in the detection stage.  

\subsection{Confidence Map}
\begin{figure}[ht] 
	\begin{center}
		\includegraphics[width=1\linewidth]{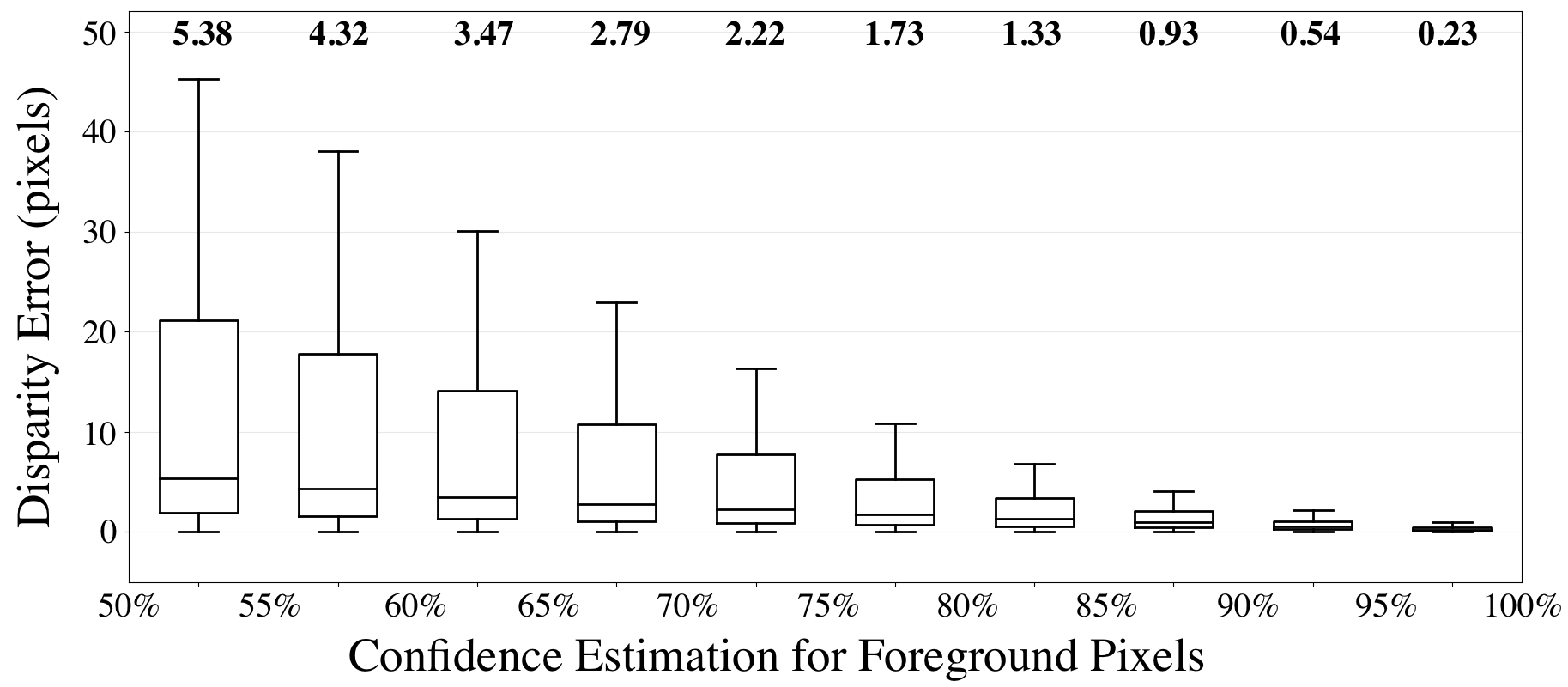}
	\end{center}
		\caption{Disparity Error vs. Confidence Estimation from modified HD$^{3}$ for all pixels belonging to cars in the \emph{validation} set on KITTI object detection benchmark~\cite{geiger2012we}. Each box represents a 5\% range in confidence. The median is shown on top of the box. There is a trend that a higher confidence estimation indicates a higher quality of the estimation output. We do not include plots with confidence lower than 50\%, because they contain significantly fewer samples.}
		\label{fig:confidence}
\end{figure}

In comparison with LiDAR point clouds, the estimated point clouds often suffer from high variance in the accuracy of depth. To consider this difference, we take advantage of the confidence estimation from the stereo matching algorithm and encode this information in the point cloud simply as an additional layer as shown in Eq.~\ref{eq:point_w_uncertainty}.
\begin{equation}\label{eq:point_w_uncertainty}
    p_c = \begin{bmatrix} x\\ y\\z\\\sigma \end{bmatrix}
\end{equation}
where $\sigma$ represents the confidence for each point.
Fig.~\ref{fig:confidence} shows the relationship between the confidence estimation and the estimation error for all the pixels belonging to cars on KITTI \emph{validation} set. The inverse relationship shown in Fig.~\ref{fig:confidence} proves that the confidence estimates indicate the accuracy of the estimation and the quality of the estimated point clouds. Adding the confidence estimation as an additional layer is shown to improve object detection performance in Section~\ref{ablation}.

\subsection{3D Box Regression Network.}
In the detection phase, we choose the open-source PointRCNN as our 3D object detector for its strong performance, and because it works directly on point clouds without voxelization, allowing us to encode the confidence estimation into the points.

\section{IMPLEMENTATION DETAILS}\label{metric}
\noindent \textbf{Semantic Segmentation.}
We employ VideoProp-LabelRelax~\cite{zhu2019improving} as the semantic segmentation network during inference. Since the KITTI semantic segmentation dataset~\cite{geiger2012we} has only 200 labelled images, the network and model was instead trained on Mapillary~\cite{neuhold2017mapillary} and Cityscapes~\cite{cordts2015cityscapes} before being finetuned on KITTI. There is a class discrepancy between the object detection benchmark and the semantic segmentation benchmark on KITTI. On the object detection benchmark, the cyclist class is a single stand-alone class, but on the semantic segmentation benchmark, the rider and bike classes are separate. To solve this issue, we dilate the rider masks and check if they overlap with a bike mask. If there is an overlap, we keep the union of the original rider mask and the bike mask as a cyclist mask. All other bike masks are discarded. \\

\noindent \textbf{Stereo Depth Estimation.} For our stereo matching algorithm, we use the model pretrained on the FlyingThings3D Dataset~\cite{mayer2016large}, and then train the proposed two-decoder network on the training split of the KITTI object detection dataset. The foreground and background decoders have the same pre-trained weights before training on KITTI. To train on KITTI, we use the depth map generated by depth completion~\cite{ku2018defense} and their corresponding point clouds as ground truth. To obtain the ground truth instance segmentation masks used during training, we follow~\cite{ku2019monocular} and project ground truth points within the 3D labels to the image as the foreground masks. Training is performed for 375 epochs, with a batch size of 32 and a learning rate of $5 \times 10^{-4}$. The learning rate decays by 0.5 at the 125\textsuperscript{th}, 187\textsuperscript{th}, and 250\textsuperscript{th} epochs. We apply horizontal flipping as data augmentation. Specifically, we increase the number of training samples by switching the left image and right image and horizontally flipping both of them. For this module, We train one model for cars, and another model for pedestrians and cyclists. \\

\noindent \textbf{3D Object Detection.}
The region proposal network of PointRCNN is trained for 200 epochs with a batch size of 16 and a learning rate of 0.001, and the 3D box refinement network is trained for 50 epochs with a batch size of 8 and a learning rate of 0.001. For augmentations, we follow the original paper of PointRCNN~\cite{shi2019pointrcnn}. Similar to PointRCNN, we subsample 16,384 points for each scene. Specifically, we sample half of the points that have depth larger than 20~m.

\section{EXPERIMENTAL RESULTS}\label{results}
We evaluate our proposed method on the widely used KITTI 3D object detection dataset~\cite{geiger2012we}. Specifically, we first compare the 3D object detection results with the state-of-the-art stereo-based detectors, then validate each contribution through ablation studies, and finally show qualitative results. KITTI contains 7,481 stereo image-pairs for training and 7,518 for testing. The benchmark also has annotations for 3 classes, which are cars, pedestrians and cyclists. Each annotation is categorized as easy, moderate, and hard based on the 2D box height, occlusion, and truncation. We follow the same training and validation split as other methods~\cite{wang2019pseudo, pon2019object, li2019stereo}. We also submit our results for all three classes to the online KITTI test server. KITTI recently changed its evaluation metrics on the test server. For the results on \emph{test} set and in the ablation studies, we use the new KITTI metric which is mean average precision with 40 recall positions. For a fair comparison with other approaches, the results on the \emph{validation} set are compared using the original KITTI metric with 11 recall positions.\\

\subsection{AP Comparison with State-of-the-Art Methods}
\begin{table*}[h!]
	\small
	\centering
	\tabcolsep=0.11cm
	\begin{tabular}{|c||c@{~/~}cc@{~/~}cc@{~/~}c||c@{~/~}cc@{~/~}cc@{~/~}c|}
		\hline
		\multirow{2}{*}{Method} & \multicolumn{6}{c||}{0.7 IoU} & \multicolumn{6}{c|}{0.5 IoU} \\
		\cline{2-13} &
		\multicolumn{2}{c}{Easy} & \multicolumn{2}{c}{Moderate} & \multicolumn{2}{c||}{Hard} &
		\multicolumn{2}{c}{Easy} & \multicolumn{2}{c}{Moderate} & \multicolumn{2}{c|}{Hard} \\ \hline
		TLNet~\cite{qin2019triangulation} & 18.15 & 29.22 & 14.26 & 21.88 & 13.72 &18.83 & 59.51 & 62.46 & 43.71& 45.99 & 37.99 & 41.92 \\
		\hline
		Stereo-RCNN~\cite{li2019stereo} & 54.11 & 68.50 & 36.69 & 48.30 & 31.07 & 41.47 & 85.84 & 87.13 & 66.28 & 74.11 & 57.24 &58.93\\
		\hline
		PL:F-PointNet~\cite{wang2019pseudo} &59.4 & 72.8 &39.8 & 51.8 & 33.5 &44.0 &89.5 & 89.8 & 75.5 & 77.6 &66.3 &68.2  \\
		\hline
		PL:AVOD~\cite{wang2019pseudo} & 61.9 & 74.9 & 45.3 & 56.8 & 39.0 & 49.0 & 88.5&89.0 & 76.4 &77.5 & 61.2 & 68.7 \\
		\hline
		PL++:AVOD~\cite{you2019pseudo} & 63.2 & 77.0 & 46.8 & 63.7 & 39.8 & 56.0 & 89.0& 89.4 &77.8 & 79.0 & 69.1 & 70.1 \\
		\hline
		PL++:PIXOR~\cite{you2019pseudo} & - & 79.7 & - & 61.1 & - & 54.5 & - & 89.9 & - & 78.4 & - &74.7\\
		\hline
		PL++:P-RCNN~\cite{you2019pseudo} & 67.9 & 82.0 & 50.1 & 64.0 & 45.3 & 57.3 & 89.7 &89.8 &78.6 & 83.8 &75.1 & 77.5\\
		\hline
		OC-Stereo~\cite{pon2019object} &64.07 &77.66 &  48.34 & 65.95 & 40.39 & 51.20 & 89.65 & 90.01 & 80.03 & 80.63 & 70.34 & 71.06 \\
		\hline
		DSGN~\cite{chen2020dsgn} &73.21 &83.24 & 54.27 & 63.91 & 47.71 & 57.83 & - & - & - & - & - & - \\
		\hline
		Ours & \textbf{76.17} & \textbf{87.31} & \textbf{57.82} & \textbf{68.69} & \textbf{54.63} & \textbf{65.80} & \textbf{90.58}  & \textbf{97.04} & \textbf{87.01} & \textbf{88.58} & \textbf{79.76} & \textbf{80.34} \\
		\hline
	\end{tabular}
	\caption{\textbf{Car Localization and Detection.} $AP_{3D}$ / $AP_{BEV}$ on KITTI \emph{validation} set. The results are evaluated using the original KITTI metric with 11 recall positions.}
	\label{tab:kitti_val_cars}
\end{table*}
\begin{table*}
	\small
	\centering
	\begin{tabular}{|c||ccc||ccc|}
		\hline
		\multirow{2}{*}{Method} & \multicolumn{3}{c||}{$AP_{3D}$} & \multicolumn{3}{c|}{$AP_{BEV}$} \\
		            &    Easy & Moderate &  Hard &  Easy & Moderate &  Hard \\ \hline
				RT3DStereo~\cite{konigshof2019realtime} & 29.90 &23.28 & 18.96 &58.81 & 46.82& 38.38
		\\ \hline
		Stereo-RCNN~\cite{li2019stereo} &   47.58 & 30.23 & 23.72 & 61.92 & 41.31 & 33.42 \\ \hline
		PL:AVOD~\cite{wang2019pseudo} & 54.53 & 34.05 & 28.25 & 67.30 & 45.00 & 38.40
		\\ \hline
		PL++:P-RCNN~\cite{you2019pseudo} & 61.11 & 42.43 & 36.99 & 78.31 & 58.01 & 51.25 \\ \hline
		OC-Stereo~\cite{pon2019object} &55.15 & 37.60 & 30.25 & 68.89 & 51.47 & 42.97 \\
		\hline
		DSGN~\cite{chen2020dsgn} &{73.50} & {52.18} & {45.14} & 82.90 & 65.05 & 56.60\\
		\hline
		Ours   & \textbf{74.39} & \textbf{53.58} & \textbf{46.50} & \textbf{85.29} & \textbf{66.44} & \textbf{58.95} \\ \hline
	\end{tabular}
	\caption{\textbf{Car Localization and Detection.} \emph{$AP_{3D}$} and \emph{$AP_{BEV}$} on KITTI \emph{test} set. The results are evaluated using the new KITTI metric with 40 recall positions. Several methods are not available on the leaderboard. }
	\label{tab:kitti_test_cars}
\end{table*}
\begin{table*}
	\small
	\centering
	
	\begin{tabular}{|c||ccc||ccc|}
		\hline
		\multirow{2}{*}{Method} & \multicolumn{3}{c||}{$AP_{3D}$} & \multicolumn{3}{c|}{$AP_{BEV}$} \\
		            &    Easy & Moderate &  Hard &  Easy & Moderate &  Hard \\ \hline
		\multicolumn{7}{|c|}{Pedestrian}\\\hline
		RT3DStereo~\cite{konigshof2019realtime} & 3.28 & 2.45 & 2.35 & 4.72 & 3.65 & 3.00
		\\ \hline
		OC-Stereo~\cite{pon2019object} & 24.48 & 17.58 &15.60 & 29.79 & 20.80 & 18.62 \\
			\hline
		DSGN~\cite{chen2020dsgn} &20.53 & 15.55 & 14.15 & 26.61 & 20.75& 18.86\\
		\hline
		Ours   & \textbf{33.22} & \textbf{24.31} & \textbf{20.95} & \textbf{39.24}  & \textbf{29.56} & \textbf{25.87} \\ \hline	\multicolumn{7}{|c|}{Cyclist}\\\hline
		RT3DStereo~\cite{konigshof2019realtime} & 5.29 & 3.37 & 2.57 & 7.03 & 4.10 & 3.88
		\\ \hline
		OC-Stereo~\cite{pon2019object} & 29.40 & 16.63 & 14.72 & 32.47 & 19.23 & 17.11 \\
		\hline
		DSGN~\cite{chen2020dsgn} &27.76 & 18.17 &16.21 & 31.23 & 21.04& 18.93\\
		\hline
		Ours   & \textbf{47.40} & \textbf{30.89} & \textbf{27.73} & \textbf{55.33} & \textbf{36.25} & \textbf{32.17} \\ \hline	
	\end{tabular}
	\caption{\textbf{Pedestrian and Cyclist Localization and Detection.} \emph{$AP_{3D}$} and \emph{$AP_{BEV}$} on KITTI \emph{test} set.  The results are evaluated using the new KITTI metric with 40 recall positions.  Several methods are not available on the leaderboard.}
	\label{tab:kitti_test_people}
\end{table*}
We compare our method with state-of-the-art stereo-based methods on the KITTI benchmark in Tables \ref{tab:kitti_val_cars}, \ref{tab:kitti_test_cars} and \ref{tab:kitti_test_people}. For the car class, our approach outperforms all state-of-the-art methods on the KITTI \emph{validation} split in all categories. On the \emph{test} set, our method ranked the first among all stereo-based methods on all three difficulties in both $AP_{3D}$ and $AP_{BEV}$. Specifically for moderate difficulty, we show a 1.40\% increase in $AP_{3D}$, and a 1.39\% increase in $AP_{BEV}$. For pedestrians and cyclists, our proposed method outperforms all other stereo-based methods by significant margins. Most noticeably, on the test server, we have 6.73\% and 12.72\% AP increase in the 3D moderate category at 0.7 IoU for pedestrians and cyclists, respectively. For classes with limited data, our decomposed pipeline allows us to pretrain the sub-modules using additional datasets, which performs significantly better than other methods that lack this ability. The total inference time of our method is 0.57s on average on a GeForce RTX 2080 Ti GPU, which is faster than the current state-of-the-art method DSGN~(0.68s)~\cite{chen2020dsgn} and is comparable with other stereo-based methods on the KITTI leaderboard~\cite{geiger2012we}.

\subsection{Ablation Studies}~\label{ablation}
We analyze the effect of each added modules in Table.~\ref{tab:ablation}. The baseline is the original HD$^3$ network with background points filtered out and PointRCNN as the 3D detector. 

For depth estimation, we follow~\cite{wang2019task} and use mean absolute relative error (absRel) and scale invariant logarithmic error (SILog) as the evaluation metrics. To better observe depth estimation accuracy improvements from the modifications made to HD3 for the more challenging but underrepresented pixels at greater depths, we present depth estimation errors for all foreground pixels with greater than 20~m depth.\\

\noindent \textbf{Effect of Split Depth Estimation.} To investigate the effect of split depth estimation, we train the modified HD$^3$ network without the point cloud loss and feed the resulting point cloud directly to the 3D detector without a confidence score layer. The separate decoders allow us to improve the depth estimation for foreground pixels by 29.8\% (from 0.047 to 0.033 absRel), and this leads to improvements in the $AP_{3D}$ by 1.81\% and $AP_{BEV}$ by 1.31\%. \\
\begin{figure*} 
	\begin{center}
		\includegraphics[width=1\linewidth]{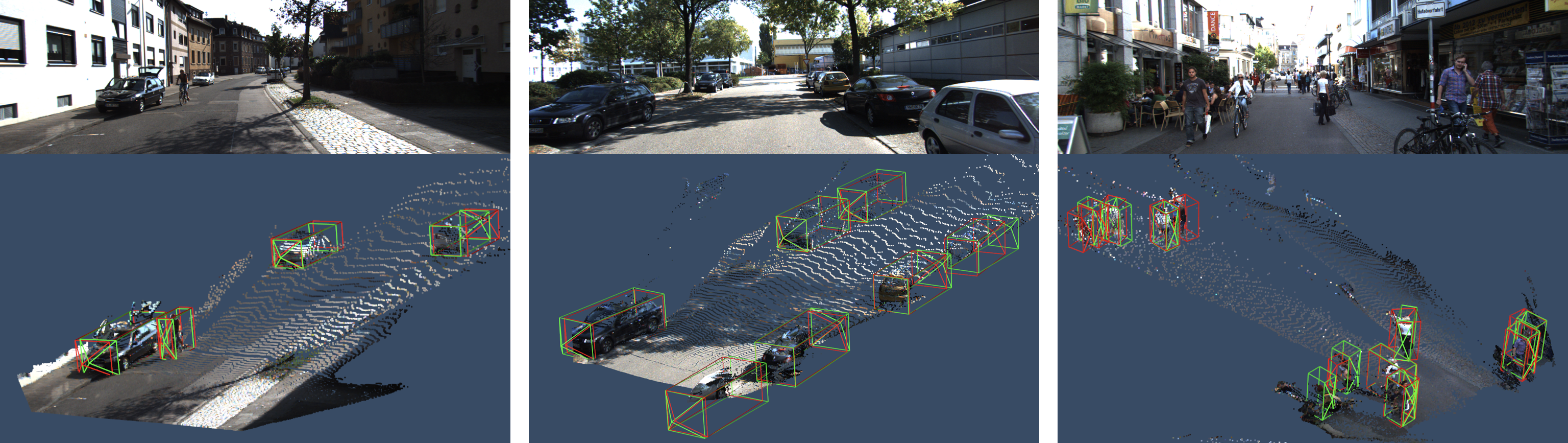}
	\end{center}
		\caption{Qualitative results of our method on several samples in the KITTI \emph{validation} split. The ground truth labels and detections are shown in red and green, respectively. }
		\label{fig:visualization}
\end{figure*}

\noindent \textbf{Effect of Point Cloud loss.}
We analyze the effect of point cloud loss by feeding the point cloud to the 3D detector without confidence estimation. Point cloud loss further improves the foreground depth estimation from 0.033 to 0.027 absRel. Including a point cloud loss in the pipeline also helps to obtain better 3D object detection accuracy, improving the $AP_{3D}$ by 2.27\% and $AP_{BEV}$ by 0.29\%.  \\

\noindent \textbf{Effect of Confidence Estimation.}
Finally, adding confidence estimation as an additional layer in the 3D detector boosts the 3D detection performance by another 1.02\% and BEV performance by another 1.73\%. This shows that the 3D detection network benefits from the confidence estimation generated by the depth estimation module.
\begin{table}[h!]
\centering
    \begin{tabular}{|c|c|c||c c|c|c|}
		\hline
		\multirow{2}{*}{\shortstack{Split \\ Depth}} &
		\multirow{2}{*}{$L_{pc}$} & 
		\multirow{2}{*}{\shortstack{Confidence \\ Feature}} &\multicolumn{2}{c|}{Foreground}&
		\multirow{2}{*}{$AP_{3D}$} & 
		\multirow{2}{*}{$AP_{BEV}$}	\\
		 & & &absRel &SILog & &\\
		\hline
		 -          & -          & -        &0.047 & 0.126 & 52.48 & 67.84 \\
         \checkmark	& -          & -        &0.033 &\textbf{0.112}  & 54.29 & 69.15 \\
         \checkmark	& \checkmark & -        & \textbf{0.027}&  \textbf{0.112}& 56.56 & 69.44 \\
         \checkmark	& \checkmark & \checkmark & -& -& \textbf{57.58} & \textbf{71.17} \\
		\hline
	\end{tabular}
   \caption{\textbf{Ablation Studies}. Comparison of depth estimation for foreground pixels, and comparisons of $AP_{3D}$ and $AP_{BEV}$ at 0.7 IoU for moderate difficulty for the car class. Both are evaluated on KITTI \emph{validation} set. $L_{pc}$ denotes the point cloud loss. $absRel$ denotes the mean absolute relative error and $SILog$ denotes the scale invariant logarithmic error.}
   \label{tab:ablation}
\end{table}

\noindent \textbf{Sensitivity to Semantic Segmentation.} The performance of our method depends on the quality of the semantic segmentation. On the KITTI benchmark, there are only 200 images with semantic ground truth for training, which limits the performance of the semantic segmentation network. To investigate the performance upper bound of our method, we perform experiments using the labels that are generated from~\cite{chen2014beat} as the segmentation masks on the KITTI \emph{validation} split. Table~\ref{tab:ablation_mask} shows that with the labelled masks, there is a 5.36\% and a 3.84\% improvement in $AP_{3D}$ and $AP_{BEV}$. This experiment suggests that our proposed method has the potential to obtain even better performance on other datasets with more accurate semantic segmentation.  

\begin{table}[h!]
\centering
    \begin{tabular}{|c||c|c|}
		\hline
		Masks & $AP_{3D}$  & $AP_{BEV}$   \\
		\hline
		VideoProp-LabelRelax~\cite{zhu2019improving} & 57.57 & 71.16 \\
		\hline
		Labels~\cite{chen2014beat}         & 62.93 & 75.00 \\
		\hline
	\end{tabular}
   \caption{Comparisons of $AP_{3D}$ and $AP_{BEV}$ for moderate difficulty for cars at 0.7 IoU using estimated segmentation masks~\cite{zhu2019improving} with using labelled masks~\cite{chen2014beat} on KITTI \emph{validation} set.}
   \label{tab:ablation_mask}
\end{table}

\subsection{Qualitative Results}
Fig.~\ref{fig:visualization} shows the estimated point cloud, the ground truth bounding box, and the final detections of our proposed method on the KITTI \emph{validation} split. The right image suggests that even with the significant improvements compared with the previous state-of-the-art methods, pedestrian and cyclist classes remain challenging for our stereo-based detector because of the limited training samples and the high intra-class variation.

\section{CONCLUSIONS}
In this paper, we present CG-Stereo, a confidence-guided stereo 3D object detection pipeline with split depth estimation. Taking advantage of the mature development of image segmentation, the stereo matching network can learn the depth for foreground and background pixels separately, and achieve better foreground depth estimation and 3D object detection performance as a result. We also show that encoding the confidence estimation from the stereo matching network into the point cloud as a soft attention mechanism guides the 3D object detector to focus more on the accurate points and further boosts 3D object detection accuracy. Our proposed method outperforms all state-of-the-art stereo-based methods on the KITTI 3D object detection benchmark. Future work includes evaluating our proposed method on other autonomous driving datasets~\cite{sun2019scalability, caesar2019nuscenes, pitropov2020canadian}. 
\newpage
{
\bibliographystyle{IEEEtran}
\bibliography{IEEEtran}
}

\end{document}